\documentclass{article}

\usepackage{microtype}
\usepackage{graphicx}
\usepackage{subfigure}
\usepackage{booktabs} 
\usepackage{amsmath}

\usepackage[hyphens]{url}

\usepackage{hyperref}

\usepackage[accepted]{icml2021}

\icmltitlerunning{Wildfire Smoke Plume Segmentation}

\begin{document}

\twocolumn[
\icmltitle{Wildfire Smoke Plume Segmentation Using\\
            Geostationary Satellite Imagery}



\icmlsetsymbol{equal}{*}

\begin{icmlauthorlist}
\icmlauthor{Jeff Wen}{stan}
\icmlauthor{Marshall Burke}{stan,fse}
\end{icmlauthorlist}

\icmlaffiliation{stan}{Earth System Science, Stanford University, Stanford, California, USA}
\icmlaffiliation{fse}{Deputy Director, Center on Food Security and the Environment}

\icmlcorrespondingauthor{Jeff Wen}{jlwen@stanford.edu}

\icmlkeywords{remote sensing, segmentation, computer vision, wildfire smoke}

\vskip 0.3in
]



\printAffiliationsAndNotice{}  

\begin{abstract}
Wildfires have increased in frequency and severity over the past two decades, especially in the Western United States. Beyond physical infrastructure damage caused by these wildfire events, researchers have increasingly identified harmful impacts of particulate matter generated by wildfire smoke on respiratory, cardiovascular, and cognitive health. This inference is difficult due to the spatial and temporal uncertainty regarding how much particulate matter is specifically attributable to wildfire smoke. One factor contributing to this challenge is the reliance on manually drawn smoke plume annotations, which are often noisy representations limited to the United States. This work uses deep convolutional neural networks to segment smoke plumes from geostationary satellite imagery. We compare the performance of predicted plume segmentations versus the noisy annotations using causal inference methods to estimate the amount of variation each explains in Environmental Protection Agency (EPA) measured surface level particulate matter $<$2.5\textmu{}$m$ in diameter ($\textrm{PM}_{2.5}$). 
\end{abstract}

\section{Introduction} 
\label{introduction}
Since the early 1980s, scientists at the National Environmental Satellite, Data and Information Service (NESDIS) have explored the use of environmental satellites to detect fires while other researchers identified the potential of using satellite imagery for wildfire smoke detection \cite{mcnamara_2004, chung_1984, svejkovsky_1985}. Although NESDIS implemented algorithmic fire detection with the introduction of the Hazard Mapping System (HMS) in 2002, wildfire smoke is still manually annotated by analysts using multi-frame animations of Geostationary Operational Environmental Satellite (GOES) imagery \cite{mcnamara_2004, ruminski_2007}. We combine the methods being developed in computer vision with geostationary satellite imagery to identify wildfire smoke plumes in near real-time, providing a method to improve analysis with more accurate identification of smoke and potentially extend analysis beyond the US.

Researchers have applied machine learning techniques to related problems with mixed results. Specifically, Wan et al. (2011) used unsupervised learning approaches to cluster smoke in RGB color images using both a sharpening and mixture model approach \cite{wan_2011}. Their exploratory results showed that it was possible to identify smoke plumes by analyzing satellite imagery, but the unsupervised approach limited opportunities to examine the out-of-sample performance of the models. Wolters and Dean present two approaches to smoke segmentation by combining labeled hyperspectral images with logistic regression models \cite{wolters_2015, wolters_2017}. Their more recent work illustrates the ability of auto-logistic regression models to capture the spatial association between neighboring pixels, which allows for smoothing over predictions. However, while the results are positive, the model's error rates are still relatively high. 

More recent literature illustrates the potential of using neural networks to estimate local pollution levels \cite{li_2017, hong_2019}. Similarly, researchers have designed custom neural network architectures for classifying different classes of aerosol including smoke, but the classification does not isolate and segment plumes, which makes it challenging to quantify the contribution wildfire smoke versus other sources of aerosol in downstream analysis \cite{ba_2019}. Filonenko et al. (2017) compares the performance of different CNN architectures for smoke detection but does not consider segmentation of plumes in the images \cite{filonenko_2017}. 

Neural networks have been successfully applied in varied image segmentation tasks such as street-level scene labeling, neural structure labeling in biomedical electron microscopy images, road network segmentation, and aerial image segmentation \cite{segnet_2015, unet_2015, mnih_2013, marmanis_2016}. These techniques have allowed applied researchers to automate previously manual, time-intensive annotation tasks with models that can be trained and adapted to different problems. 

Ramasubramanian et al. (2019) use a convolutional neural network architecture with 6 bands of GOES-16 imagery as input and HMS satellite-derived annotations that were subsequently corrected by experts as labels \cite{ramasubramanian_2019}. Their model uses a single timestamp as input and relies on subject matter experts to manually correct the imperfect labels, which resulted in a dataset size of approximately 120 scenes containing smoke. The authors note that their model performs smoke pixel classification using the pixel of interest in addition to an input neighborhood around the pixel of interest. This approach results in predicted boundaries that extend beyond the visible smoke boundary and relies on a manually defined input-pixel neighborhood size. This along with the need for expert corrected labels makes it difficult to scale the approach to further analysis.

Larsen et al. (2021) utilize a fully convolutional network (FCN) to identify wildfire smoke plumes from satellite observations obtained by the Himawari-8 geostationary satellite situated over East Asia and Western Pacific regions \cite{larsen_2021}. They use a deterministic cloud-masking algorithm to generate smoke versus non-smoke pixel labels, which are then used to train the FCN. While the approach is similar, the manually annotated smoke plumes in our work may better capture variability in the visual representation of smoke compared to a deterministic algorithm. 

Other recent literature similarly use deep convolutional neural networks (CNNs) for identifying smoke plumes from UAV and drone images as well as synthetically generated smoke images \cite{frizzi_2021, yuan_2019}. These works show promising performance but differ from the work presented here as the aerial or synthetic images used provide a side-view angle of smoke rather than a top-down view. Li et al. (2018) and Zhu et al. (2020) both use 3D convolution based CNN approaches to segment smoke plumes from videos, but these video sequences also present a side-viewing angle \cite{li_2018, zhu_2020}. While the side-view images may provide higher resolution, their availability can be inconsistent compared to geostationary satellite observations. Given the temporal frequency of geostationary satellite imagery, future work can explore using video segmentation networks on sequences of geostationary images.

\begin{figure}[ht]
\vskip 0.1in
\begin{center}
\centerline{\includegraphics[width=\columnwidth]{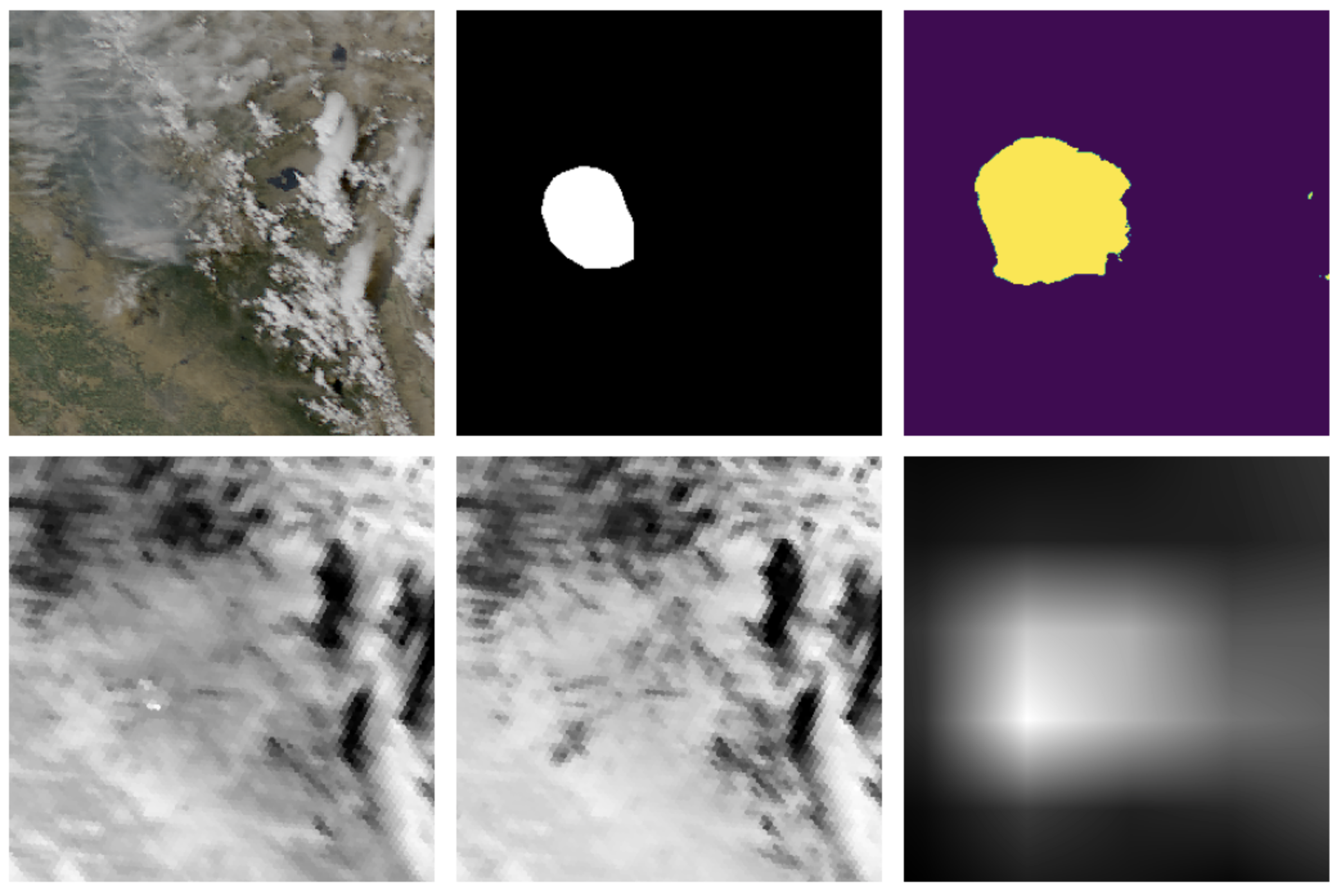}}
\caption{First row: Model input, ground truth annotation label, and model predicted segmentation after thresholding. Second row: Additional input channel 07, channel 11, and MERRA-2.}
\label{fig:pred_ex}
\end{center}
\vskip -0.3in
\end{figure}

In this work, we adapt CNNs to automatically identify wildfire smoke plumes from satellite imagery. We use satellite observations as input and HMS annotated smoke plumes as the target labels to train our models. Because analysts often generate these plume annotations using multiple hours of satellite imagery, we investigate different methods for improving training with these noisy labels. To validate our models, we compare the predicted smoke plume segmentations against the HMS annotated smoke plumes on PM2.5 measurements from EPA monitoring stations. Better understanding of the spatial extent of wildfire smoke is especially important in downstream research, such as in the environmental economics literature, where accurate identification of smoke plumes could lead to better estimates of the causal effect of smoke exposure.

\section{Methodology}
\subsection{Data}
\label{data}

The main imagery data used are satellite earth observations captured by the Geostationary Operational Environmental Satellite (GOES-16) positioned over the eastern United States. GOES-16 along with the west coast GOES-17 satellite provide frequent (every 5 minutes for the Continental US) multi-spectral measurements that are largely used for weather modeling and storm tracking over the Atlantic and Pacific oceans respectively. Images are downloaded if there are associated smoke plumes during 2018 from the HMS annotations. In order to ensure that the imagery captures radiance from the earth's surface, images were limited to between 12PM and 11PM Coordinated Universal Time (UTC). Furthermore, due to the large number of daily observations, we limited training images to California and Nevada from the 2018 wildfire season between May-December of 2018. However, we perform testing on unseen data between May-October of 2019 and 2020. For testing on external data from 2019-2020, we download 3 images per day between May-October. These multiple images allow us to identify if multiple plumes are overhead on a given day. Given the frequency of GOES imagery, in future work we hope to increase the temporal resolution of the plumes.

\begin{table}[t]
\caption{Input channels}
\label{tab:input_channels}
\vskip 0.1in
\begin{center}
\small\addtolength{\tabcolsep}{-3pt}
\begin{small}
\begin{sc}
\begin{tabular}{lccc}
\toprule
Name & Wavelength & Type & Primary uses \\
\midrule
Blue      & 0.47 $\mu$m & Visible  & Aerosols \\
Red       & 0.64 $\mu$m & Visible  & Clouds \& ash \\
Veggie    & 0.86 $\mu$m & Near-IR  & Vegetation \\
Shortwave & 3.9 $\mu$m  & IR & Fire hotspots \\
Cloud-Top & 8.4 $\mu$m  & IR & Cloud-top \\
\bottomrule
\end{tabular}
\end{sc}
\end{small}
\end{center}
\vskip -0.1in
\end{table}

After collecting the smoke plume data and the associated GOES-16 imagery from the Amazon Web Services (AWS) NOAA open data repository, the raw imagery is reprojected and a pseudo true-color composite is generated from the red, blue, and "green" bands using the SatPy package \cite{satpy}. Although the GOES-16 satellite carries multiple sensors, it lacks a green band, which must be generated to create a true-color composite \cite{bah_2018}. Additionally, the 07 and 11 channels are also included in experiments as these near-infrared channels can be used for fire hotspot detection and cloud-top identification (examples in Figure \ref{fig:pred_ex}). Each of the downloaded images was saved as a 1200x1200 \texttt{.geotiff} image for a total of 615 images. These were then randomly cropped to generate up to 15 different 300x300 images where 60\% of the crops had a smoke plume to deal with class imbalance. As a result, there were 6825 images which was split into 70\% training, 15\% validation, and 15\% testing. We further ensured that the cropped images generated from the same base image could only be in one set of data to reduce potential data leakage.

We pair the satellite imagery with smoke plume labels from the HMS archives. While the labels extend back to 2006, we use observations from the most recent generation of geostationary satellite (GOES-16) launched in mid-2017 as this subset of data allows for more spectral bands to be used in the analysis. Analysts often use multi-hour animations to draw the extent of the smoke plumes, which presents challenges as the extent over time might not match the boundaries of smoke plumes in a single image due to wind or other meteorological factors \cite{brey_2018}. Although the current labels are manually annotated from satellite imagery, one goal of this research is to identify the extent to which we can learn meaningful segmentations from these imprecise labels. We attempt to improve training on noisy labels by providing additional signal using the Modern-Era Retrospective analysis for Research and Applications, Version 2 (MERRA-2) aersol optical thickness (AOT) measurements of particulate matter as an additional input channel \cite{gelaro_2017}. The MERRA-2 data combines multiple sources of aerosol optical density information mainly derived from the Moderate Resolution Imaging Spectroradiometer (MODIS) instrument aboard NASA's Terra and Aqua satellites. We discuss 2 additional methods to account for the noisy labels in the next section.

In the rest of the paper, we will refer to the true-color image as 1 band and the true-color, channel 07, and channel 11 combined as 3 bands. Lastly, 4 bands refers to using the true-color, channel 07, channel 11, and the MERRA-2 in the input.

\subsection{Model}
\label{model}
Our baseline experiments utilize an adapted U-Net neural network architecture, originally designed for biomedical image segmentation, to segment smoke plumes from a pseudo true-color RGB image \cite{unet_2015}. The adapted network keeps the same number of layers in the encoding and upsampling blocks but decreases the number of convolutional filters. Furthermore, we replace the ReLU activation function with PReLU activations for improved stability in training. We also include multiple spectral bands in subsequent experiments as these bands (shown in Table \ref{tab:input_channels}) provide contextual information of fire hotspot location and cloud-top phase in channel 07 and 11 respectively. 

Across the model training, we use the Adam optimization with a learning rate starting at $\alpha$=5e-5 and stepping down by $\gamma$=0.1 every 9 epochs \cite{kingma_2017}. Each model was trained for a total of 21 epochs using a batch size of 16. We compared binary cross entropy (BCE) and mean absolute error (MAE) losses during the training and validation process. To track accuracy during training and validation, we kept track of the average Dice coefficient.

\begin{equation}
\text{Dice Coefficient}=\frac{2|A \cap B|}{|A|+|B|}
\label{eq:dice}
\end{equation}

This statistical similarity metric first proposed by Lee Raymond Dice in 1945 and shown in Equation \ref{eq:dice} calculates 2 times the amount of overlapping pixels between the predicted ($A$) and ground truth mask ($B$) divided by the total number of pixels in both masks \cite{dice_1945}. 

Additionally, we consider two methods for explicitly handling noisy labels in the training process. Specifically, we experiment with using mean absolute value of error (MAE) loss, which has been shown to be tolerant to label noise, and data sampling to specify that only low loss training samples contribute to the gradient updates \cite{ghosh_2017, xue_2019}. The loss sampling strategy allows the model to run the prediction on a batch of inputs as usual, but before calculating the average loss across samples, we identify the training sample in the batch that produced the highest loss. Then, we set the loss for that example to 0 so that it does not contribute to making weight updates. As mentioned by Xue et al., this approach assumes that as the model performance improves, particularly noisy samples can result in high loss, which can have large impact on the weight updates. This training strategy mitigates the effect of these samples.

\begin{figure}[ht]
\vskip 0.1in
\begin{center}
\centerline{\includegraphics[width=\columnwidth]{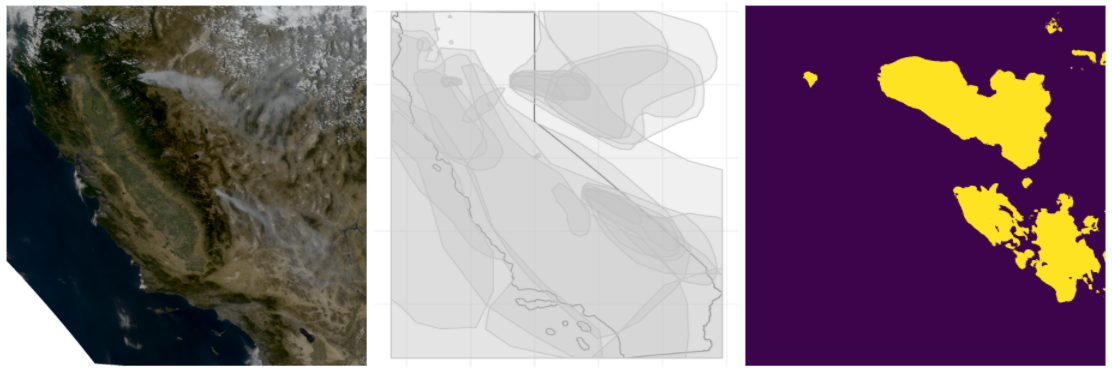}}
\caption{The model predicted segmentation on the right more closely matches the visible smoke in the input imagery compared to the HMS annotated smoke plumes in the middle, which cover most of California. The input on the left is a test image from September 8th, 2019 that was not used for training or validation.}
\label{fig:test_pred_ex}
\end{center}
\vskip -0.3in
\end{figure}

\subsection{External validation}
\label{external_validation}

While the DICE coefficient is used for validation during training, we apply the models on unseen images from 2019 and 2020 to estimate the performance of these models compared to the hand annotated HMS smoke plumes from the same years. This allows us to test the models in a setting most similar to downstream inference tasks where the models would be used to identify smoke plumes across a certain time frame, then used for additional analysis. 

Specifically, we leverage econometric tools for causally identifying the contribution of wildfire smoke to changes in ground level PM2.5 readings as measured by the EPA. We use a quasi-experimental approach to exploit variation in smoke and PM2.5 over time to estimate the effect of identified wildfire smoke plumes on PM2.5 readings. We use fixed effect panel regressions (shown in Equation \ref{eq:fe_reg}) with station fixed effects to account for time-invariant unobserved effects such as the fact that different stations may have unobserved characteristics correlated with smoke exposure. We set the station PM2.5 reading as the dependent variable and consider the HMS annotations as well as different model predicted smoke plumes as the independent variable.

\begin{equation}
\text{PM2.5}_{it}=\beta_{1} \text{Smoke}_{it}+\alpha_{i}+\epsilon_{it}
\label{eq:fe_reg}
\end{equation}

This approach allows us to compare performance even in the presence of noisy annotation labels by measuring performance against an external ground truth. In the above equation, the $\alpha_{i}$ captures the EPA station fixed effects, which would handle time-invariant station level differences. The $\text{Smoke}_{it}$ variable is determined by the presence or absence of smoke plumes overlapping an EPA station where the smoke plumes come from either the HMS annotations or model predicted segmentations. 

\begin{table}[t]
\caption{Model performance on validation set. Bold rows represent the models that achieve the best validation DICE coefficient for a specific loss function. The asterisk denotes that the model weights were updated after removing the highest loss sample per batch.}
\label{tab:model_perf}
\vskip 0.1in
\begin{center}
\small\addtolength{\tabcolsep}{-3pt}
\begin{small}
\begin{sc}
\begin{tabular}{lccc}
\toprule
Bands & Loss & Avg. Loss & DICE \\
\midrule
1           & BCE           & 0.2535            & 0.0948          \\
\textbf{3}  & \textbf{BCE}  & \textbf{0.2236}   & \textbf{0.1074} \\
3*          & BCE           & 0.2313            & 0.1008          \\
4           & BCE           & 0.1884            & 0.1028          \\
1           & MAE           & 0.0986            & 0.2635          \\
3           & MAE           & 0.0986            & 0.2649          \\
\textbf{3*} & \textbf{MAE}  & \textbf{0.0986}   & \textbf{0.2655} \\
\textbf{4}  & \textbf{MAE}  & \textbf{0.0986}   & \textbf{0.2655} \\
\bottomrule
\end{tabular}
\end{sc}
\end{small}
\end{center}
\vskip -0.1in
\end{table}

To compare the HMS annotated smoke plumes against the model predicted segmentations, we consider the adjusted R2 (\texttt{Adj.R2}) and within-adjusted R2 (\texttt{W Adj.R2}) metrics. The \texttt{Adj.R2} metric measures the total amount of variation in PM2.5 measured at EPA monitoring stations explained by the smoke plumes and the \texttt{W Adj.R2} metric represents the amount of variation in PM2.5 that is explained "within" the PM2.5 station unit by the different sources of smoke data. This distinction is important and we prefer the "within" metric because it is a better indicator of ability to explain variation in PM2.5 using the different smoke data as it only considers remaining variation separate from time-invariant unobserved station level differences. 

\section{Results}

\begin{figure}[ht]
\vskip 0.1in
\begin{center}
\centerline{\includegraphics[width=\columnwidth]{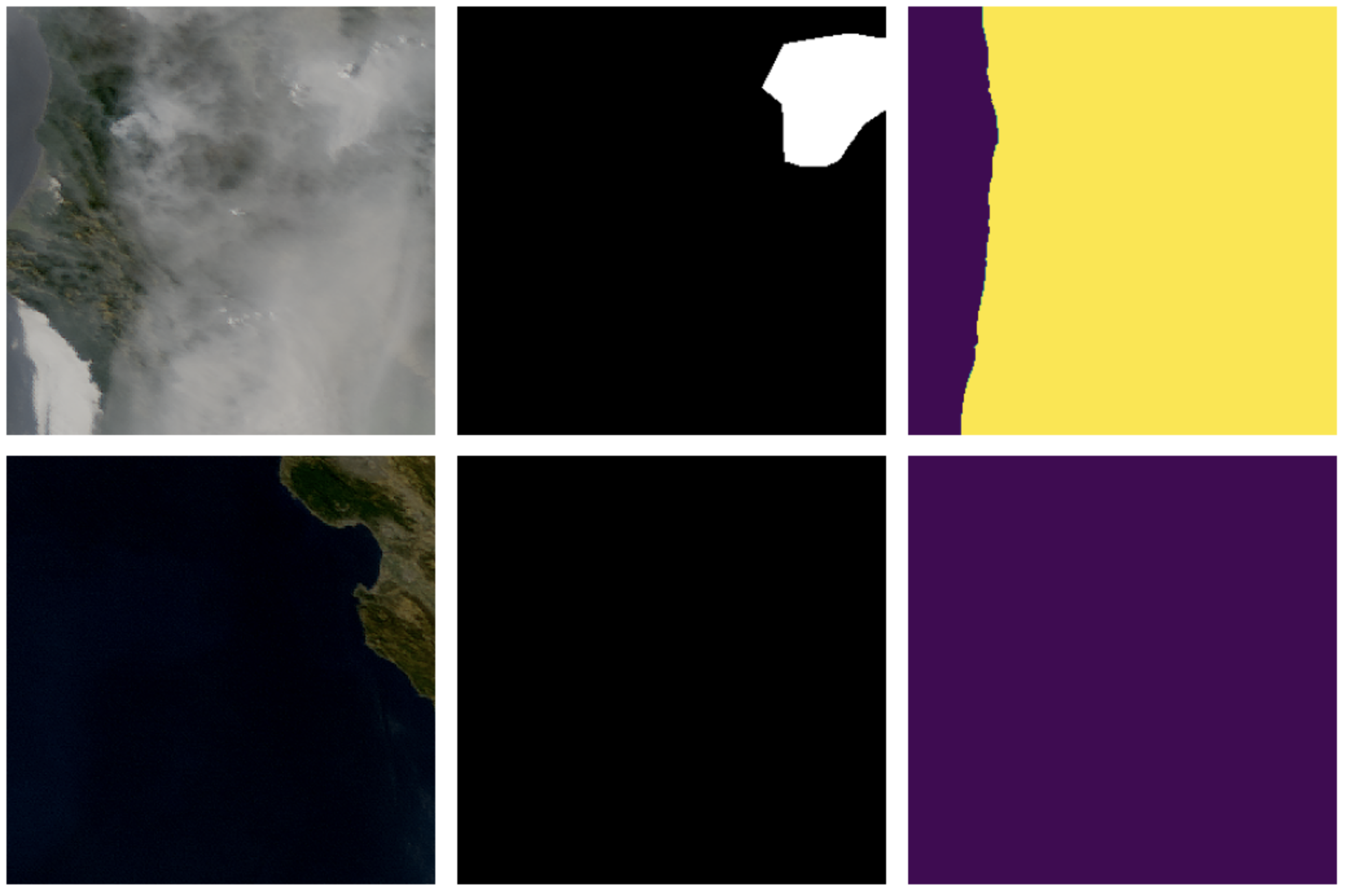}}
\caption{Additional examples show model predictions on unseen test images with a lot of smoke in the first row and no smoke in the second row. Model input, ground truth annotation label, and model predictions are displayed in both rows from left to right.}
\label{fig:add_ex}
\end{center}
\vskip -0.3in
\end{figure}

Our preliminary results show that qualitatively the BCE loss models are able to learn to segment smoke from the input imagery. The top right panel in Figure \ref{fig:pred_ex} shows the performance of the model on a validation sample and the rightmost panel in Figure \ref{fig:test_pred_ex} and Figure \ref{fig:add_ex} show segmentations of the model trained with BCE loss and 3 input channels on previously unseen test images. The segmentations appear to more precisely locate the smoke in the input imagery on the left compared to the annotations in the middle. Even though there is cloud cover in Figure \ref{fig:pred_ex}, the model is able to learn to differentiate between the cloud cover and the smoke plumes. Further quantitative results are shown in Table \ref{tab:model_perf}, where the bolded rows represent the best validation DICE coefficient when using either of the loss functions during training. The asterisk denotes that the model weights were updated only after removing training samples that had the highest loss per batch \cite{xue_2019}. 

Although the MAE loss resulted in higher overall accuracy, the qualitative results indicated that these models were predicting "no-smoke" for nearly all pixels and were unable to deal with the label imbalance in the images. Even with the addition of the additional step to sample only low loss training samples, the model performance with MAE loss was still qualitatively unable to identify smoke pixels compared to the models trained with BCE loss. 

\begin{table}[t]
\caption{HMS annotated vs. model predicted plume performance on EPA PM2.5 measurements. Different sources of smoke plumes are validated against external PM2.5 measurements using a fixed effect panel regression approach to compare variation explained. The asterisk denotes the loss sampling model results.}
\label{tab:fe_perf}
\vskip 0.1in
\begin{center}
\small\addtolength{\tabcolsep}{-3pt}
\begin{small}
\begin{sc}
\begin{tabular}{lcccccc}
\toprule
 & Adj. R2 & W Adj. R2 \\
\midrule
Annotations & 0.2316 & 0.1946    \\
1 band & 0.2707 & 0.2367    \\
3 band & 0.2715 & 0.2374    \\
4 band & 0.3038 & 0.2703    \\
3 band* & 0.2987 & 0.2653    \\ 
\bottomrule
\end{tabular}
\end{sc}
\end{small}
\end{center}
\vskip -0.1in
\end{table}

When we evaluate the model performance on previously unseen data and compare the performance against the annotations, we see that the model predictions are able to explain more of the variation in surface level PM2.5 readings regardless of model specification (results shown in Table \ref{tab:fe_perf}). The higher values for the model predicted smoke data suggest that while these results are preliminary, the CNN segmented smoke plumes may explain more of the changes in PM2.5 compared to the hand annotated smoke plumes. It is important to note that the hand annotations are meant to capture visible smoke, while the particulate matter captured at EPA stations might not be visible from satellite imagery. Therefore, the models leveraging the MERRA-2 channel as additional input might not be a fair comparison against the hand annotations, which are mainly generated using visible imagery. However, as seen in Table \ref{tab:fe_perf}, even the model trained only using RGB as input appears to capture more of the "within" station variation than the hand annotations.

While in Table \ref{tab:model_perf} the model trained by removing the influence of the highest loss sample per batch appeared to perform worse than the best 3-band model using BCE loss, in the external validation (shown in Table \ref{tab:fe_perf}), this model was able to explain more of the total and within variation in PM2.5. Furthermore, this model was nearly able to match the performance of the 4-band model illustrating that explicitly ignoring high loss training samples when there are noisy labels could be beneficial for downstream tasks. This might not be apparent from the validation accuracy metrics because the validation dataset also includes noisy labels, which make it challenging to gauge model performance. This underscores the importance of clean validation data especially in the presence of noisy labels.

\section{Conclusion}
As wildfires continue to worsen, it is increasingly important to quantify the effects of wildfire smoke exposure on society. In this work, we used an adapted U-Net architecture to segment smoke plumes from geostationary satellite imagery with the goal of improving our understanding of the spatial extent of smoke. We further leveraged quasi-experimental methods to compare the variation in EPA station PM2.5 measurements that could be explained by either model predicted segmentations or hand annotated plumes. While smoke plumes have been manually annotated in the United States since the 2000s, our results suggest that automated segmentation methods are at least qualitatively comparable to the annotated smoke plumes and explain more of the within station PM2.5 variation in external validation. These early results show the potential of adapting neural networks for improving downstream inference even with noisy labels. 

In future work, we look to focus on additional methods to improve the model robustness to noisy labels. We also hope to extend and develop these methods to identify smoke across the globe and over time to better inform the impacts of wildfires.

\subsection*{Software and Data}
The Geostationary Operational Environmental Satellite (GOES) data used in this work is available for download at \url{https://registry.opendata.aws/noaa-goes}. Hazard Mapping System smoke data is available at \url{https://www.ospo.noaa.gov/Products/land}. The models were developed using PyTorch version 1.7.1 and the fixed effects estimators were estimated using the Fixest R package version 0.8.4 \cite{pytorch, fixest}. The models and data will be made available at the time of publication.

\subsection*{Acknowledgements}
We thank reviewers whose comments and suggestions helped clarify and improve this manuscript. We also thank the Sustainability and Artificial Intelligence Lab for discussions.

\bibliography{main}
\bibliographystyle{icml2021}

\end{document}